\documentclass{article}

\PassOptionsToPackage{numbers, compress}{natbib}
\usepackage[preprint]{neurips_2026}

\usepackage[utf8]{inputenc} 
\usepackage[T1]{fontenc}    
\usepackage{hyperref}       
\usepackage{url}            
\usepackage{booktabs}       
\usepackage{amsfonts}       
\usepackage{nicefrac}       
\usepackage{microtype}      
\usepackage{xcolor}         

\usepackage{amsmath}
\usepackage{algorithm}
\usepackage{algpseudocode}
\usepackage{graphicx}
\usepackage{makecell}

\title{CORE: Contrastive Reflection Enables Rapid Improvements in Reasoning}

\author{%
  Linas Nasvytis\thanks{Equal contribution.} \\
  Stanford University \\
  \And
  Simon Jerome Han\footnotemark[1] \\
  Stanford University \\
  \And
  Ben Prystawski \\
  Stanford University \\
  \AND
  Satchel Grant \\
  Stanford University \\
  \And
  Noah D. Goodman \\
  Stanford University \\
  \And
  Judith E. Fan \\
  Stanford University \\
}

\begin{document}

\maketitle

\begingroup
\renewcommand{\thefootnote}{}
\footnotetext{%
\parbox{\dimexpr\linewidth-\footnotesep\relax}{%
Correspondence: \texttt{linasmn@stanford.edu}\\ 
Code is available at: \url{https://github.com/LinasNas/core-reasoning}
}}
\endgroup
\setcounter{footnote}{0}

\begin{abstract}

Language models can use verifiable rewards to improve at a wide variety of reasoning tasks. However, both parametric (e.g. RLVR) and non-parametric (e.g. prompt optimization) approaches to doing so typically require hundreds of training samples and thousands of model rollouts, making them expensive in the best case and intractable in the worst. To address this challenge, we introduce Contrastive Reflection (CORE), a non-parametric learning algorithm that compares past reasoning traces to generate {\it insights}: short natural-language descriptions of reasoning strategies and constraints that capture differences between successful and unsuccessful problem attempts. 
Across four reasoning tasks, we demonstrate that CORE enables more rapid improvement than both parametric (GRPO) and non-parametric (GEPA, episodic RAG, and MemRL) methods, while using fewer rollouts. 
Under fixed rollout budgets with as few as five training samples, CORE achieves the strongest performance in most task–data regimes.
Finally, we highlight how CORE is substantially more context-efficient than non-parametric baselines, requiring fewer prompt tokens while storing learned knowledge as compact, interpretable natural-language insights. 
Our results therefore suggest that distilling contrasts between successful and unsuccessful reasoning traces into abstract and useful insights can provide a more efficient and interpretable route to model self-improvement than weight updates, prompt optimization, or direct reuse of stored reasoning traces.

\end{abstract}
\section{Introduction}

Language models can learn from verifiable rewards, but often require large amounts of data and compute to do so. For example, parametric methods such as GRPO can require hundreds of thousands of rollouts for a model to make meaningful progress on a given task \citep{guo2025deepseek}, while non-parametric methods such as GEPA use fewer rollouts but still rely on hundreds of training and validation samples to make comparable gains \citep{agrawal2025gepa}. Humans, by contrast, can often improve substantially at a new task with only a handful of practice problems and learning trials \citep{nam2024systematic}. What accounts for this difference, and how might we enable language models to learn from verifiable rewards with more human-like efficiency?

Here, we consider {\it insight discovery} as one potential answer to these questions. A rich body of work in cognitive psychology suggests that rapid learning in humans depends partly on the ability to revisit past successes and failures in order to discover more abstract, explicit, and concise principles that explain their difference \citep{cushman2020rationalization,schwartz1998time,alfieri2013learning}. Comparing what has worked to what has not is crucial, as is estimating how useful these learned insights are: people often acquire more general and reusable insights when contrasting past experiences, rather than reflecting upon them in isolation \citep{schwartz1998time,alfieri2013learning}, and people often apply insights selectively when solving new problems based on what is currently relevant or was previously useful \citep{lisman2005hippocampal,adcock2006reward}.
Lastly, we take high-level inspiration from theories of how the human brain makes use of multiple memory systems, including one system for encoding specific experiences and a complementary system for distilling more general knowledge from these experiences \citep{mcclelland1995there}.
Although prior work has shown that language models can improve beyond scalar rewards alone with the use of verbal reflections or text-based feedback \citep{shinn2023reflexion, song2026expanding,shi2026experiential}, these contrastive and utility-based mechanisms that can consolidate prior experience into more abstract, reusable knowledge are largely absent from existing approaches to learning from verifiable rewards. 


To that end, we introduce Contrastive Reflection (CORE), a non-parametric learning algorithm that enables a frozen language model to learn from verifiable rewards with both greater sample and rollout efficiency than existing methods. CORE works by building two external memory stores: one that stores generated insights (`insight memory'), and one that stores past rollouts (`rollout memory'). With every failed problem attempt during training, CORE prompts the language model to contrastively reflect by comparing pairs of its own successful and unsuccessful past rollouts in order to generate insights about the strategies and constraints that might capture differences between them. These insights are then selectively retrieved and placed in-context when the model encounters future problems on the basis of both semantic similarity and utility estimates that are updated with each new associated success and failure. CORE therefore differs from existing approaches to learning from verifiable rewards in what it stores: transparent, natural-language insights paired with utility estimates, rather than weight updates, individual rollouts, or a single global prompt.

\begin{figure}[t]
  \centering
  \includegraphics[width=0.7\linewidth]{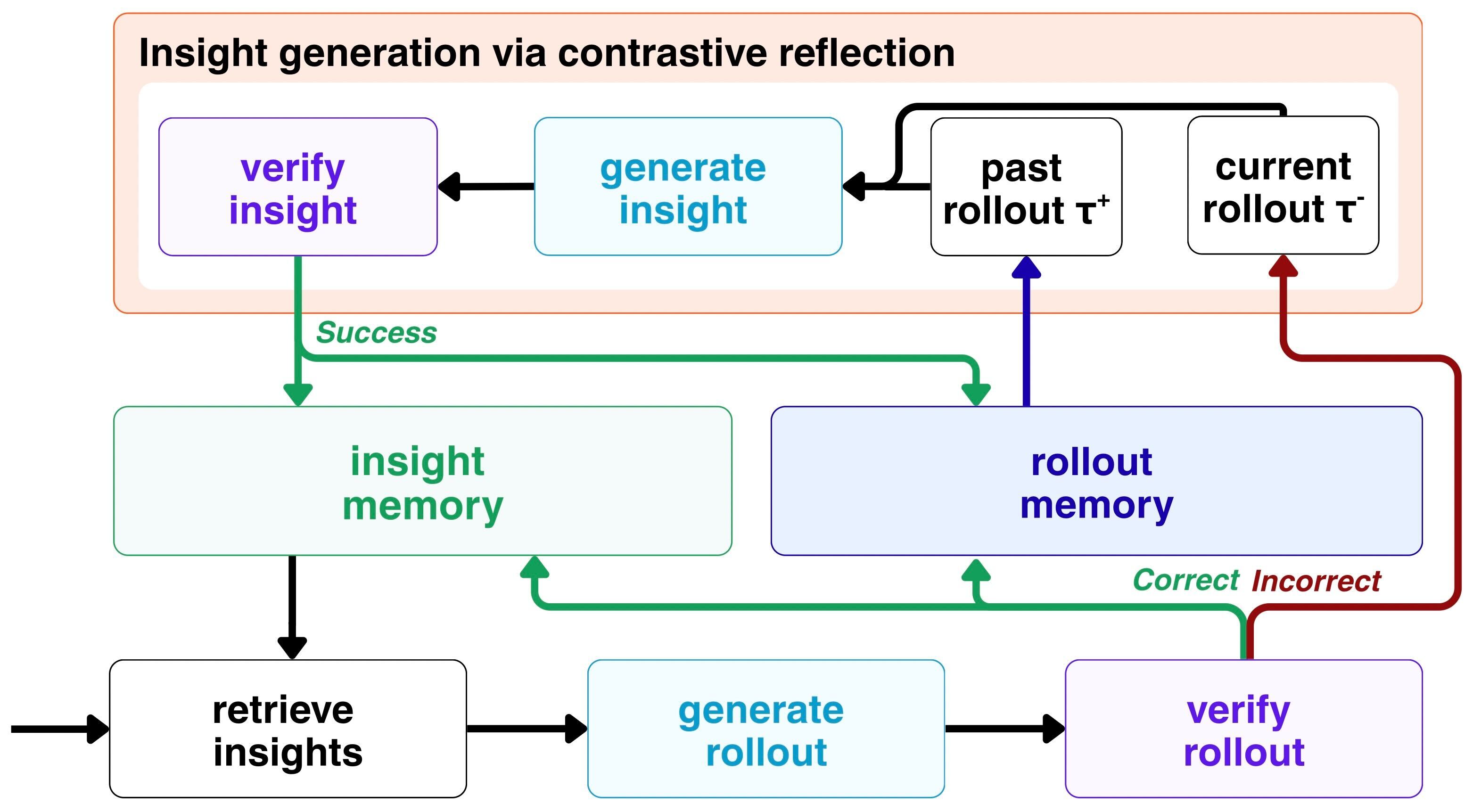}
  \caption{Illustration of the CORE algorithm: A model retrieves relevant insights, then generates a reasoning trace and answer to a question. If the model does not answer correctly, we generate new insights by contrasting the failed reasoning trace against a similar problem that the model answered correctly. Insights that lead the model to solve the problem correctly are added to the memory store.}
  \label{fig2}
\end{figure}

Our contributions are as follows:


\begin{enumerate}
    \item We introduce CORE, a non-parametric learning algorithm that enables language models to learn from verifiable rewards by generating and accumulating insights.
    \item Across a range of logic, planning, and problem-solving tasks, we demonstrate that CORE can outperform competing parametric and non-parametric baselines while learning faster and adding substantially less evaluation-time context, regardless of the number of available training samples.
    \item We demonstrate that CORE insights are {\it interpretable} learning artifacts expressed in natural language and associated with empirical utility estimates, reducing the risk of introducing unwanted behaviors from opaque parameter updates.
    \item We identify the components for generating useful insights through ablations that demonstrate how contrastive reflection and utility-aware retrieval each contribute to performance.
\end{enumerate}

\section{Related Work}

{\bf Learning from Verifiable Rewards.} 
Existing methods for learning from verifiable rewards can be classified as parametric or non-parametric. Parametric methods update model weights, for example by training on idealized rationales (STaR \citep{zelikman2022star}) or by rewarding successful attempts via automatic verifiers (GRPO \citep{cobbe2021training,shao2024deepseekmath,guo2025deepseek}). Non-parametric methods keep the model frozen and instead improve its context, for example by refining a single task prompt with task metrics (MIPRO) or textual feedback (GEPA) \citep{opsahl2024optimizing,agrawal2025gepa}. Both types of approaches typically encode what is learned in forms that are either hard to interpret or hard to reuse selectively. In contrast to these methods, CORE keeps the model frozen but stores what is learned as a set of natural-language insights that can be retrieved selectively per problem and inspected after the fact.

{\bf External Memory Systems for Language Models.} Current external memory systems for language models typically vary along two dimensions that affect sample efficiency: what exactly is stored, and how items are selected for retrieval. Stored items range from raw experience -- full reasoning traces and verbal reflections on task feedback \citep{shinn2023reflexion} -- to increasingly abstracted forms: executable programs \citep{wang2023voyager}, distilled behavior descriptions \citep{didolkar2025metacognitive, ouyang2025reasoningbank}, and human-authored procedural guidance such as \texttt{AGENTS.md} \citep{xu2026agent}. Retrieval is typically based on semantic similarity, with MemRL extending this by learning utility scores over episodic memories \citep{zhang2026memrl}. However, recent work shows that continuously consolidating trajectories into textual memory can be unstable, sometimes even degrading performance below no-memory baselines -- motivating memory systems that curate \textit{which} raw episodes are used for consolidation and \textit{when} consolidation occurs, rather than updating textual memory after every interaction \citep{zhang2026useful}. CORE differs from these approaches along both dimensions. It stores natural-language insights derived from curated contrasts between failed attempts and successful attempts on semantically similar problems, triggers this abstraction step only after failures, and admits candidate insights only when they improve performance on the originating problem. It retrieves these insights using both semantic similarity and each insight's measured utility on related problems.

{\bf Learning Efficiency.} Recent methods for language-model self-improvement still require substantial compute and data for training, and gains along one dimension of learning efficiency often come at the cost of another. Two dimensions matter: \emph{sample efficiency}, the number of distinct training problems needed for learning, and \emph{rollout efficiency}, the number of attempts needed for those problems. Among parametric methods, Reinforcement Learning from Text Feedback (RLTF) adds natural-language critiques to scalar rewards to improve rollout efficiency, but its largest reasoning setting still trains on 19{,}800 Reasoning Gym examples \citep{song2026expanding}. Experiential Reinforcement Learning (ERL) improves rollout efficiency by adding a gated reflection-and-retry stage for sparse and delayed feedback, but still operates in large-data regimes, using 10{,}000 procedurally sampled training instances for FrozenLake and Sokoban and 4 samples per prompt for each attempt in its compute-matched RLVR comparison \citep{shi2026experiential}. Process-supervision methods can improve rollout efficiency by providing denser step-level feedback, but require expensive step-level labels or reliable process reward models \citep{lightman2023let}. Among non-parametric methods, GEPA improves rollout efficiency over GRPO on some tasks, but still uses up to 7{,}000 rollouts per task on reported benchmarks \citep{agrawal2025gepa}. CORE addresses both efficiency dimensions: by extracting explicit reusable insights from each scored attempt and later assigning each of them credit for improved or diminished performance, it can match or exceed these baselines using fewer training problems and fewer rollouts.
\section{CORE: Contrastive Reflection}

CORE is a non-parametric learning algorithm that allows a frozen language model to improve using explicit, contrastive, and utility-aware reflection. The central learned object in CORE is an {\it insight}: a short natural-language description of a general reasoning strategy or constraint that can help with future problems. Insights are not summaries of prior rollouts; instead, they are better viewed as credit-assignment hypotheses about what distinguishes successful rollouts from unsuccessful ones. CORE produces insights by contrastively reflecting about pairs of past rollouts, tests them by using verifier feedback, stores them for future use only when they improve model performance, and retrieves them on the basis of their continuously updated utility estimates. Figure~\ref{fig2} illustrates a single training step for this algorithm, and pseudocode can be found in Appendix~\ref{app:pseudocode}.

Beyond the frozen language model $M$ itself, CORE's implementation relies on four components: an external memory store, failure-biased problem sampling, insight retrieval, and contrastive reflection.

\paragraph{Problem setting.}
Let $\mathcal D_{\mathrm{train}}$ and $\mathcal D_{\mathrm{eval}}$ denote training and held-out evaluation problem sets for a given task. We consider tasks where each problem $q \in \mathcal D_{\mathrm{train}} \cup \mathcal D_{\mathrm{eval}}$ is associated with an existing verifier $V_q$ that maps a candidate answer to a binary reward $r \in \{0,1\}$. The solver is a language model $M$ that remains frozen throughout. For each training problem $q$ we maintain a baseline success-rate estimate $b_q$ which captures the expected performance of $M$ on $q$ without learning updates. $b_q$ is initialized from $n_{\mathrm{base}}$ independent samples and updated when additional observations are collected absent learning updates. This allows the baseline-relative utility of any learning update to be defined as
\[
U(q) = r - b_q.
\]
This normalization gives positive evidence only when an update-conditioned attempt improves over the baseline for that problem, preventing inflation from easy problems.


\paragraph{External memory store.} CORE maintains two external memory stores. The `rollout memory' $\mathcal R$ stores correctly solved past rollouts. A rollout
\[
\tau = (q, I, y, r)
\]
contains a problem $q$, the retrieved set of insights $I$, the model output $y$, and the verifier reward $r$. Each rollout is indexed by an embedding of its problem. Rollout memory is used to find semantically similar past problems and construct pairs of rollouts for contrastive reflection. The insight memory $\mathcal I$ stores short natural-language insights together with empirical evidence about where each insight has been useful. For each problem $p$ on which insight $i$ has been applied, CORE stores the number of applications $N_{p,i}$ and the empirical mean baseline-relative utility $\bar U_{p,i}$ for that insight--problem pair.

\paragraph{Failure-biased problem sampling.} Because CORE only generates new insights after failures, we bias our problem sampling towards problems that the model solves less reliably. For training problem $q_i$, let $n_i^{\mathrm{base}}$ be the number of baseline samples used to estimate $b_i$, $n_i^{\mathrm{epi}}$ be the number of training attempts on $q_i$, and $c_i^{\mathrm{epi}}$ be the number of correct attempts. CORE estimates current accuracy as
\[
a_i =
\frac{
n_i^{\mathrm{base}} b_i + c_i^{\mathrm{epi}}
}{
n_i^{\mathrm{base}} + n_i^{\mathrm{epi}}
},
\]
and assigns sampling weight
\[
w_i = 1 - a_i
\]

to $q_i$. To maintain coverage of the training set, CORE mixes this failure-biased distribution with uniform sampling. If $N = |\mathcal D_{\mathrm{train}}|$, then:

\[
p_i =
(1-\epsilon_{\mathrm{mix}})
\frac{w_i}{\sum_{j=1}^{N} w_j}
+
\epsilon_{\mathrm{mix}} \frac{1}{N}.
\]

Each problem therefore retains a minimum sampling probability of $\epsilon_{\mathrm{mix}}/N$.

\paragraph{Insight retrieval.} Given a sampled problem $q$, CORE retrieves insights in two steps. First, CORE retrieves a broad set of previously encountered training problems that are semantically similar to $q$. Let $\mathcal P(\mathcal E)$ denote the set of unique training problems indexed in rollout memory. We embed $q$ as $e_q$ and define $\mathcal N_q$ as the set of $Z$ nearest training problems under cosine similarity:
\[
\mathcal N_q =
\left\{p_1,\ldots,p_Z\right\},
\qquad
s(e_q,e_{p_1}) \geq \cdots \geq s(e_q,e_{p_Z}),
\]
where $s(e_q,e_p)$ denotes cosine similarity between problem embeddings. Then, CORE scores each insight by aggregating its observed utilities over this neighborhood:
\[
\hat U(q,i)=
\frac{
\sum_{p \in \mathcal N_q : (p,i)\text{ observed}}
N_{p,i}\,\bar U_{p,i}
}{
\sum_{p \in \mathcal N_q : (p,i)\text{ observed}}
N_{p,i} + \lambda
}.
\]
Weighting by $N_{p,i}$ gives more influence to problem--insight pairs with more observations. If insight $i$ has not been applied to any problem in $\mathcal N_q$, its local utility estimate is set to zero before exploration bonuses are applied. Exploration bonuses are defined as
\[
B(i)=
\beta \sqrt{\frac{\log(T+1)}{N_{\mathrm{global}}(i)+1}},
\]
where $N_{\mathrm{global}}(i)$ is the total number of retrievals for insight $i$ during training, $T$ is the total number of retrievals for all insights during training, and $\beta$ controls the strength of exploration. The retrieval score during training is thus
\[
S_{\mathrm{train}}(q,i)=\hat U(q,i)+B(i).
\]
CORE uses this score to retrieve the top-$K$ insights and appends them to the solver prompt for $M$. $M$ then attempts to solve $q$, and the verifier returns a reward $r$ that CORE uses to update the utility score of each retrieved insight $i \in I$ using the baseline-relative utility $U(q)$. Specifically,
\[
N_{q,i} \leftarrow N_{q,i}+1,
\]
\[
\bar U_{q,i} \leftarrow \bar U_{q,i}
+ \frac{U(q)-\bar U_{q,i}}{N_{q,i}}.
\]
CORE also increments $N_{\mathrm{global}}(i)$ and $T$, and stores the newly generated rollout in rollout memory. This update assigns the same observed utility to every insight retrieved for a solve, yielding a group-level credit assignment rule. CORE addresses this limitation through its admission procedure: new insights are tested individually before entering insight memory, while later retrieval updates estimate where admitted insights are useful across many problems.

\paragraph{Contrastive reflection.} When CORE fails to correctly solve a problem during training, it triggers a contrastive reflection step. Here, the failed attempt becomes the negative rollout $\tau^-$. To obtain a positive rollout $\tau^+$, CORE retrieves the most similar correct rollout from rollout memory, as determined by problem embeddings. Importantly, $\tau^+$ can be for the same problem as $\tau^-$ if that problem has been correctly solved before. If rollout memory has multiple rollouts for the selected problem, then CORE will choose the one with the most similar reasoning trace embedding. If rollout memory does not yet have any correct rollouts, then CORE will skip contrastive reflection altogether.

Given $(\tau^-, \tau^+)$, the same frozen model $M$ is prompted to generate a small set of candidate insights. Here, the contrastive structure is important: a rollout may be successful due to subtle reasons and unsuccessful due to many reasons. By retrieving $\tau^-$ and $\tau^+$ that are semantically similar, CORE therefore makes it easier for the model to surface insights that are meaningful. 

CORE filters out duplicate and pre-existing candidate insights and then admission-tests the remaining candidates before they are added to the insight memory. During admission-testing, CORE prompts $M$ with each candidate $i$ as the only in-context insight and samples $n_{\mathrm{adm}}$ solutions for the originating problem $q$. With $\hat a_{q,i}$ denoting the success rate, CORE admits $i$ into insight memory only if
\[
\hat a_{q,i} > b_q + \delta,
\]
where $\delta \geq 0$ is an admission margin. In our experiments, we use $n_{\mathrm{adm}}=1$ and $\delta=0$, so a candidate insight is admitted if using it in-context leads the model to solve the problem on the first attempt. Admitted insights are initialized in insight memory with the utility scores that were observed during their admission trial, and rejected insights are discarded.

\paragraph{Evaluation.} During evaluation, both rollout and insight memory are frozen. For each held-out problem $q \in \mathcal D_{\mathrm{eval}}$, CORE retrieves the top-$K$ insights using the exploitation-only score
\[
S_{\mathrm{eval}}(q,i)=\hat U(q,i),
\]
with no exploration bonus, no reflection, no admission testing, and no memory updates. Test time retrieval uses only training problems stored in rollout memory, and test time rollouts are never added to either memory.

\begin{figure}[t]
  \centering
  \includegraphics[width=\linewidth]{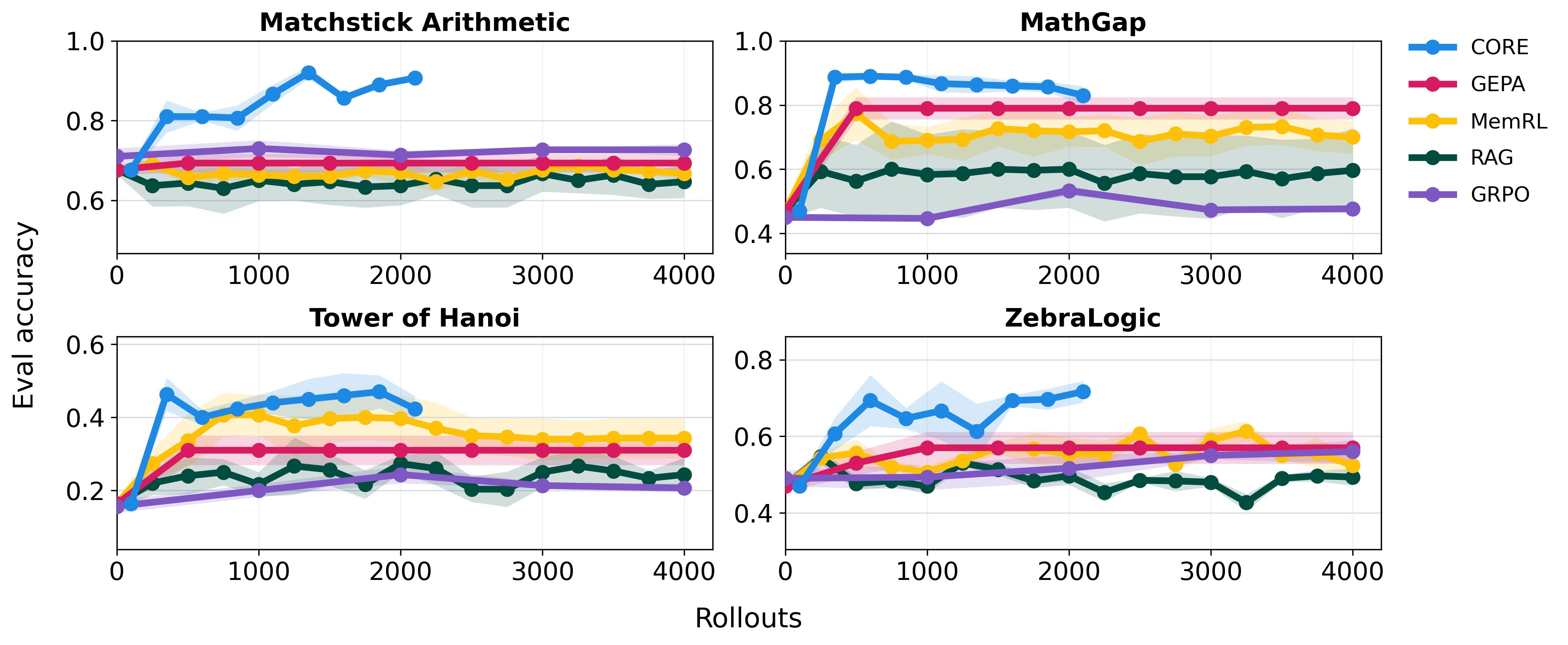}
  \caption{\textbf{CORE improves rollout efficiency.} Held-out evaluation performance as a function of total training rollouts in the 10-example regime. Across all four tasks, CORE learns more quickly and more effectively than GEPA, MemRL, RAG, and GRPO.}
  \label{fig:rollout_efficiency}
\end{figure}

\section{Evaluation}
\subsection{Setup}
\textbf{Tasks.} We evaluate CORE on four verifiable reasoning tasks spanning algorithmic, arithmetic, logical, and symbolic problem solving. We selected these tasks because they remain unsaturated for OpenAI \texttt{gpt-oss-120b}, the open-source reasoning model used in our experiments, whose performance on standard benchmarks for reasoning performance approaches proprietary models \citep{agarwal2025gpt}.  In each task, solving one example can reveal reusable skills or strategies that could transfer to some (but not necessarily all) other problems from the same domain. {\bf Tower of Hanoi} requires generating valid action sequences for a classic planning problem with recursive structure \citep{shojaee2025illusion}. {\bf MathGAP} consists of arithmetic word problems with controllable proof structure and complexity \citep{opedal2024mathgap}. {\bf ZebraLogic} includes logic-grid puzzles from constraint-satisfaction problems with controllable search complexity \citep{lin2025zebralogic}. {\bf Matchstick arithmetic} consists of matchstick equation puzzles, a classic domain in the study of human insight problem solving \citep{knoblich1999constraint,ollinger2008investigating}, with specific problem types adopted from \citep{nasvytis2026leveraging}. Each puzzle presents an invalid arithmetic equation composed of Roman numerals and operators that are rendered as matchsticks, and the model is tasked with moving a single stick to make the equation valid. We implement a problem generator and verifier for matchstick arithmetic and include the code and datasets in our \href{https://github.com/LinasNas/core-reasoning}{GitHub repository}.

\textbf{Training.} For each of the above tasks, we run evaluations on CORE and our baseline algorithms using training sets of 5, 10, and 100 problems. In all regimes, held-out performance is measured on a separate evaluation set of 100 problems. This design allows us to measure how well each algorithm can support robust learning given small training sets. For each task, method, and training-set size, we run three independent training runs and report mean held-out verifier accuracy.


\textbf{Model.} We use OpenAI \texttt{gpt-oss-120b} \citep{agarwal2025gpt} as the frozen language model in all experiments and for all learning algorithms. Unless otherwise noted, generation uses temperature $0.6$, maximum output length $32{,}768$ tokens and reasoning effort set to \texttt{high}. For MemRL, following the authors' recommendations, memory-building and adjustment calls use temperatures $0.0$ and $0.3$, respectively, and memory retrieval uses OpenAI \texttt{text-embedding-3-large} embeddings. For CORE and Episodic RAG text embeddings, we use \texttt{jina-embeddings-v2-base-en} \citep{gunther2023jina}, a 137M parameter BERT-based embedding model. For inference, we use NVIDIA NIM and Cerebras API services, matching all model-facing parameters across backends and verifying that baseline held-out accuracies were comparable across all tasks. For CORE, we retrieve top-$K_{\mathrm{train}} = K_{\mathrm{eval}} = 25$ insights per task. We use \texttt{gpt-oss-120b} for two reasons: first, to demonstrate that CORE works for larger model sizes, and second, because our initial experiments had suggested that CORE generates more useful insights with larger model sizes.

\textbf{Baselines.} We compare CORE against three non-parametric baselines and one parametric baseline, chosen to represent strong alternative approaches to learning from verifiable rewards: {\bf GRPO} \citep{guo2025deepseek} to represent parametric RLVR, {\bf GEPA} \citep{agrawal2025gepa} to represent state-of-the-art prompt optimization, {\bf Episodic RAG} to represent retrieving entire past rollouts, and {\bf MemRL} \citep{zhang2026memrl} to represent value-aware episodic memory, where past rollouts are summarized into shorter experiences and the retrieval of such experiences takes into account their past value. Our GEPA implementation uses the standard DSPy implementation, and our MemRL implementation is adapted from the official codebase and ported to our single-turn task interface, using the strongest public configuration without task-specific tuning on the evaluation set. Our RAG implementation is a setting where all rollouts are added to a rollout memory and indexed according to their problem embedding. For any model inference during training and testing, we retrieve the most similar past successful rollout and the most similar past unsuccessful rollout to place in-context alongside their verifier feedback. If the rollout memory does not yet contain any successful or unsuccessful rollouts, then we retrieve similar rollouts without accounting for success. Finally, our GRPO implementation uses LoRA via the Tinker API \citep{tml2025tinker} with a rank of 32, a learning rate of $1e-5$, a batch size of 10, and a group size of 10. Because our focus is rapid learning under limited example and rollout budgets, GRPO results should be interpreted as a rollout-limited RLVR comparison rather than a fully scaled RL training run.
\subsection{CORE achieves stronger performance with fewer rollouts}

\begin{table*}[t]
\centering
\footnotesize
\setlength{\tabcolsep}{3.2pt}
\begin{tabular}{lccc ccc}
\toprule
& \multicolumn{3}{c}{\textbf{Matchstick Arithmetic}} & \multicolumn{3}{c}{\textbf{MathGAP}} \\
\cmidrule(lr){2-4}\cmidrule(lr){5-7}
Method & 5 train items & 10 train items & 100 train items & 5 train items & 10 train items & 100 train items \\
\midrule
No Learning & 0.681 $\pm$ 0.007 & 0.681 $\pm$ 0.007 & 0.681 $\pm$ 0.007 & 0.472 $\pm$ 0.008 & 0.472 $\pm$ 0.008 & 0.472 $\pm$ 0.008 \\
GRPO & 0.630 $\pm$ 0.020 & 0.637 $\pm$ 0.022 & 0.590 $\pm$ 0.006 & 0.393 $\pm$ 0.022 & 0.400 $\pm$ 0.023 & 0.443 $\pm$ 0.012 \\
GEPA & 0.687 $\pm$ 0.019 & 0.693 $\pm$ 0.023 & 0.770 $\pm$ 0.079 & 0.853 $\pm$ 0.027 & 0.790 $\pm$ 0.035 & 0.777 $\pm$ 0.003 \\
Episodic RAG & 0.703 $\pm$ 0.066 & 0.627 $\pm$ 0.041 & 0.640 $\pm$ 0.038 & 0.770 $\pm$ 0.035 & 0.590 $\pm$ 0.126 & 0.710 $\pm$ 0.084 \\
MemRL & 0.700 $\pm$ 0.015 & 0.647 $\pm$ 0.012 & 0.703 $\pm$ 0.050 & 0.747 $\pm$ 0.043 & 0.713 $\pm$ 0.062 & 0.833 $\pm$ 0.015 \\
CORE (ours) & \textbf{0.873 $\pm$ 0.023} & \textbf{0.907 $\pm$ 0.003} & \textbf{0.870 $\pm$ 0.010} & \textbf{0.873 $\pm$ 0.009} & \textbf{0.830 $\pm$ 0.029} & \textbf{0.843 $\pm$ 0.009} \\
\addlinespace[1.0em]
\midrule[\heavyrulewidth]
\addlinespace[0.5em]
& \multicolumn{3}{c}{\textbf{Tower of Hanoi}} & \multicolumn{3}{c}{\textbf{ZebraLogic}} \\
\cmidrule(lr){2-4}\cmidrule(lr){5-7}
Method & 5 train items & 10 train items & 100 train items & 5 train items & 10 train items & 100 train items \\
\midrule
No Learning & 0.179 $\pm$ 0.007 & 0.179 $\pm$ 0.007 & 0.179 $\pm$ 0.007 & 0.509 $\pm$ 0.014 & 0.509 $\pm$ 0.014 & 0.509 $\pm$ 0.014 \\
GRPO & 0.077 $\pm$ 0.009 & 0.120 $\pm$ 0.006 & 0.107 $\pm$ 0.027 & 0.523 $\pm$ 0.027 & 0.533 $\pm$ 0.027 & 0.520 $\pm$ 0.010 \\
GEPA & 0.433 $\pm$ 0.103 & 0.310 $\pm$ 0.040 & 0.353 $\pm$ 0.185 & 0.597 $\pm$ 0.068 & 0.570 $\pm$ 0.042 & \textbf{0.707 $\pm$ 0.015} \\
Episodic RAG & 0.287 $\pm$ 0.003 & 0.243 $\pm$ 0.050 & 0.303 $\pm$ 0.048 & 0.540 $\pm$ 0.015 & 0.493 $\pm$ 0.032 & 0.497 $\pm$ 0.015 \\
MemRL & \textbf{0.517 $\pm$ 0.069} & 0.393 $\pm$ 0.047 & \textbf{0.727 $\pm$ 0.100} & 0.683 $\pm$ 0.012 & 0.543 $\pm$ 0.052 & 0.587 $\pm$ 0.020 \\
CORE (ours) & 0.400 $\pm$ 0.049 & \textbf{0.423 $\pm$ 0.035} & 0.427 $\pm$ 0.058 & \textbf{0.700 $\pm$ 0.036} & \textbf{0.717 $\pm$ 0.028} & 0.663 $\pm$ 0.022 \\
\bottomrule
\end{tabular}
\caption{\textbf{CORE improves sample efficiency}. Final held-out accuracy across training-example regimes. Entries show mean $\pm$ SEM across completed runs. The no-learning baseline is averaged across rollout-0 evaluations from all training-size regimes. Bold indicates the highest mean accuracy for each task and training-size regime.}
\label{tab:sample_efficiency_combined}
\end{table*}

We first evaluate CORE's {\it rollout efficiency}, the number of rollouts needed to achieve meaningful task improvement, in the 10-sample training regime. For each method, we measure held-out accuracy as a function of the number of training rollouts. To ensure a fair comparison, we count CORE's initial baseline accuracy runs and insight admission tests as rollouts. Results are displayed in Figure~\ref{fig:rollout_efficiency}. 

Across all four tasks, CORE learns rapidly: by 350 training rollouts (first evaluation), it already exceeds the best evaluation performance achieved by any baseline method at any training point. Despite the fact that we run each of our baselines for 4000 training rollouts and CORE for only 2100, we find that CORE also settles on higher final performance than any baseline. 

Averaged across tasks, CORE's held-out accuracy improves by $59.9\%$ from rollout 0 to rollout 350, increasing from $0.445$ to $0.712$, and sustains the gains until rollout 2{,}100\footnote{For CORE, rollout counts include the initial 100 rollouts used to estimate per-problem baseline accuracies, so the first evaluation occurs at 350 training rollouts and the final evaluation occurs at 2100 rollouts.}, achieving a $0.717$ held-out accuracy. For individual tasks, CORE improves from rollout 0 to rollout 2100 by $34.5\%$ on Matchstick Arithmetic, $76.6\%$ on MathGAP, $159.2\%$ on Tower of Hanoi, and $50.0\%$ on ZebraLogic. These results suggest that CORE can learn from a small number of training samples more quickly and more effectively than any of the baseline methods. 

\begin{figure}[t]
  \centering
\includegraphics[width=0.9\linewidth]{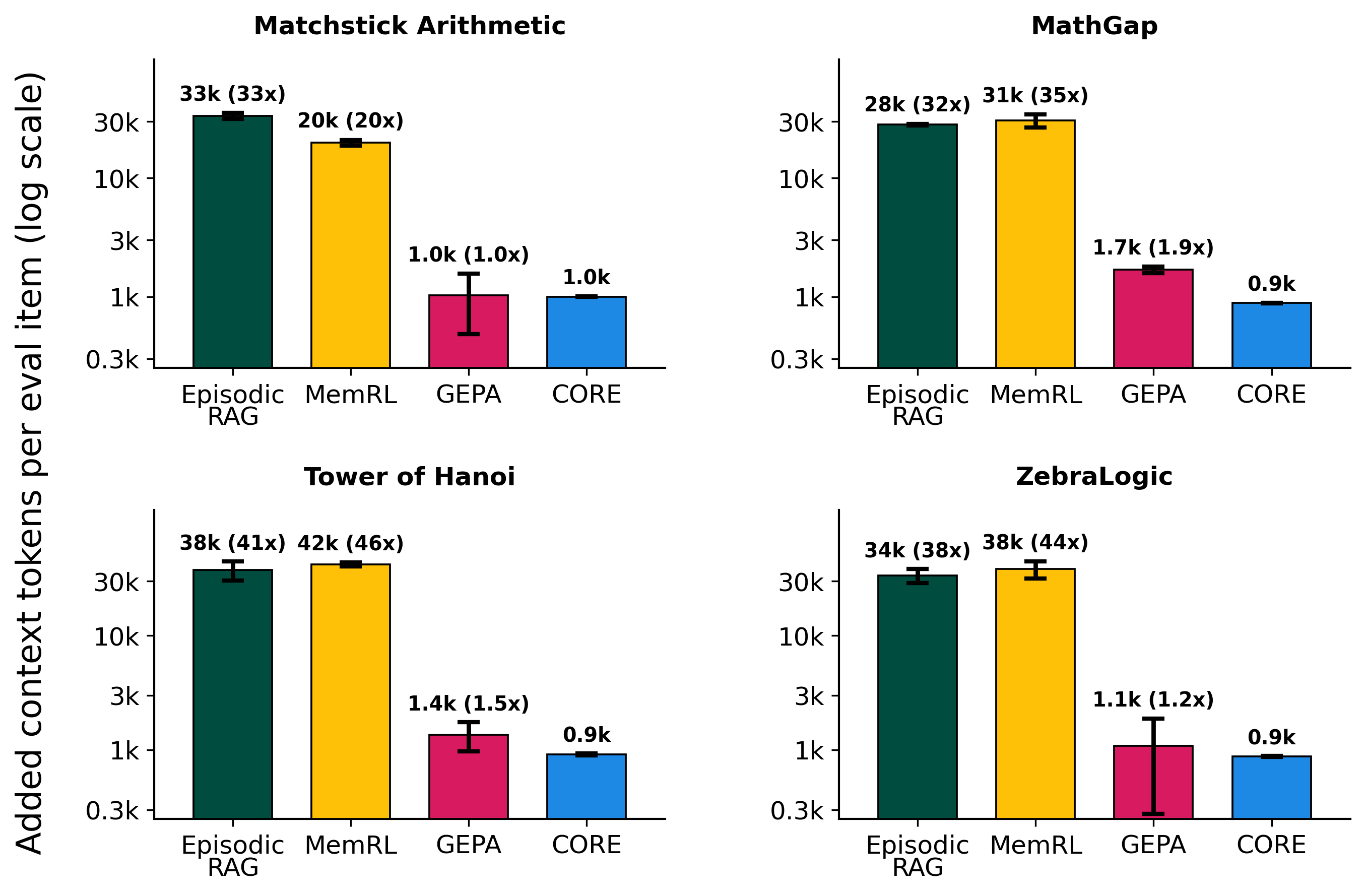}
\caption{\textbf{CORE improves context efficiency.} Added task-external context tokens per evaluation item for each method (lower number denotes higher efficiency). Error bars show SEM across training-data regimes. Labels above bars report mean added tokens and the fold increase relative to CORE. \textbf{The y-axis is log-scaled}.}
\label{fig:context_efficiency}
\end{figure}

\subsection{CORE achieves stronger performance with fewer training samples}

We next evaluate CORE's {\it sample efficiency} across different training data regimes. Specifically, we compare CORE against our baselines under fixed rollout budgets while varying the number of training samples. Table~\ref{tab:sample_efficiency_combined} illustrates held-out evaluation performance when each method is trained with 5 training samples for 2050 rollouts, 10 samples for 2100 rollouts, and 100 samples for 3000 rollouts\footnote{In settings with different numbers of training samples, the initial number of rollouts that CORE uses to calculate per-problem baseline accuracies is calculated as number of training samples multiplied by ten. To give our baseline methods a better chance, we allow them to train for these additional rollouts as well, which is why we report results for 2050, 2100, and 3000 rollouts. For all data regimes, we only run CORE for 2000 actual training rollouts.}.

CORE achieves the highest mean held-out accuracy in 9 of 12 task-by-data-regime conditions. Averaged across tasks, CORE improves over the no-learning baseline by $54.8\%$, $56.2\%$, and $52.3\%$ with 5, 10, and 100 training examples, respectively. The only conditions in which CORE does not achieve the highest mean held-out accuracy across all other baselines are Tower of Hanoi with 5 and 100 training samples (where MemRL achieves the highest accuracy), and ZebraLogic with 100 training samples (where GEPA achieves the highest accuracy). Overall, these results suggest that CORE can extract insights from training sets of different sizes across diverse reasoning tasks.

\subsection{CORE adds significantly less evaluation-time context}

We also evaluate {\it context efficiency}: how many additional tokens each non-parametric method and CORE adds to the evaluation prompt, beyond the base task prompt and answer-format instructions. Figure~\ref{fig:context_efficiency} shows the average added context per evaluation item, averaged across the 5-, 10-, and 100-sample training regimes.


CORE is the most context-efficient method we evaluate. Averaged across tasks, CORE adds $0.92$k tokens per evaluation item, compared with $33.6$k for Episodic RAG and $32.7$k for MemRL -- each $36.6\times$ and $35.6\times$ more, respectively. 
CORE is about 1.4× more context-efficient than GEPA, a meta-prompt optimization baseline, which adds an average of $1.29$k tokens per evaluation item.

Thus, CORE's gains do not come from placing large retrieved traces into the evaluation prompt; instead, it compresses training-time experience into a small set of abstract, reusable insights. Taken together, these findings suggest that CORE not only outperforms baseline algorithms across the vast majority of task and data regimes, but does so while adding substantially less evaluation-time context.

\subsection{CORE prioritizes insights that capture recurring reasoning patterns}

\begin{table}[t]
\centering
\footnotesize
\begin{tabular}{p{0.18\linewidth} p{0.28\linewidth} p{0.46\linewidth}}
\toprule
\textbf{Insight type} & \textbf{What it captures} & \textbf{Example learned insight} \\
\midrule

\parbox[t]{\linewidth}{\raggedright\textbf{Search space}\\\textbf{structuring}}
&
Identifies broad constraints, search heuristics, or specific reusable solution patterns that reduce the number of possibilities to consider.
&
\textit{Matchstick Arithmetic:} ``Use chain equality: converting a minus into a second \texttt{=} can create a three-part equality, allowing the equation to be solved when the three terms become identical.'' (Utility: 0.14)
\\

\midrule

\parbox[t]{\linewidth}{\raggedright\textbf{Intermediate state}\\\textbf{tracking}}
&
Keeps intermediate steps, quantities, or moves updated as each step changes the problem.
&
\textit{MathGAP:} ``When a clue describes a transfer or split, update the giver's and all receivers' counts immediately, and keep pre- and post-transfer values separate for later use.'' (Utility: 0.09)

\\

\midrule

\parbox[t]{\linewidth}{\raggedright\textbf{Verification and}\\\textbf{validation}}
&
Checks candidate solutions against task constraints, detects contradictions, backtracks, or satisfies output requirements.
&
\textit{Tower of Hanoi:} ``Before finalizing your answer, simulate each move step-by-step, confirming that the move obeys the one-disk, top-disk, and size-order rules, and record the resulting peg states.'' (Utility: 0.11)

\\

\bottomrule
\end{tabular}
\caption{Examples of high-utility insights learned by CORE, grouped by functional role. Utilities denote the empirical baseline-relative utility associated with each insight.}
\label{tab:insight_types}
\end{table}

Finally, we highlight that CORE stores learning artifacts as compact natural-language insights in insight memory. To characterize these artifacts, we examine one run from each task in the 10-training-example regime. We analyze how CORE's memory grows, how insight utility is distributed, and what types of insights it contains.

\textbf{Memory growth.} Averaging across all CORE runs, after 2{,}000 rollouts the number of admitted insights was highest for Matchstick Arithmetic ($143$), followed by ZebraLogic ($126$), Tower of Hanoi ($119$), and MathGAP ($65$).

\textbf{Utility distribution.} Insight utility is concentrated unevenly across memory (Figure~\ref{fig:utility_histograms}). Most admitted insights have non-negative estimated utility, with non-negative weighted mass above $91\%$ for all tasks. The distributions are approximately unimodal for Matchstick Arithmetic and ZebraLogic, but bimodal for MathGAP and Tower of Hanoi. This pattern suggests that CORE admits many mildly useful insights, while a smaller subset of high-utility insights drives the largest gains.

\textbf{Analysis of insights.}
To understand what CORE learns, we manually inspected high-utility insights from each task and grouped near-duplicates by functional role (Table~\ref{tab:insight_types}). Insights fell into three broad categories: \emph{search-space structuring}, which includes both broad heuristics and specific reusable solution patterns; \emph{intermediate-state tracking}, which keeps quantities, assignments, moves, or equations updated as reasoning unfolds; and \emph{verification and validation}, which checks constraints, detects contradictions, and enforces output requirements. These patterns suggest that CORE stores procedural abstractions for guiding future reasoning, rather than summaries of prior episodes.

\section{What Accounts for Learning from Contrastive Reflection?}

We use ablations to test which components of CORE drive its learning gains. We evaluate these ablations in the 10-example setting across all four tasks. The ablations ask two questions: whether contrastive reflection improves the generation of useful insights, and whether empirical utility estimates improve their retrieval.

\textbf{Does contrastive reflection improve insight generation?} CORE generates insights by comparing a failed reasoning trace with a successful one from the same or a similar problem. We compare this with two non-contrastive variants: one that reflects only on the most recent incorrect trace (\textit{incorrect trace only}), and one that reflects only on a correct trace (\textit{correct trace only}). Aggregating across all four tasks, CORE achieves the highest improvement over rollout 0 at every nonzero evaluation checkpoint, reaching a final mean improvement of $0.268$, compared with $0.234$ for incorrect-only reflection and $0.203$ for correct-only reflection. This suggests that the strongest insights come from explicitly comparing failure and success, rather than reflecting on either trajectory in isolation.

\textbf{Does utility-aware retrieval improve insight reuse?}
CORE retrieves insights using both semantic relevance and learned utility estimates. We compare this with a variant that retrieves insights using only the semantic similarity between the problems to which an insight has been applied and the current problem. Removing utility-aware retrieval reduces the final mean improvement from $0.268$ to $0.227$, indicating that relevance alone is not sufficient for best performance.

All CORE variants outperform GEPA, the strongest baseline, in final held-out accuracy, but full CORE achieves the largest gains. This suggests that reusable insight generation is itself a strong learning mechanism, and that it performs best with both contrastive reflection and utility-aware reuse.

\begin{figure}[t]
  \centering
  \includegraphics[width=0.95\linewidth]{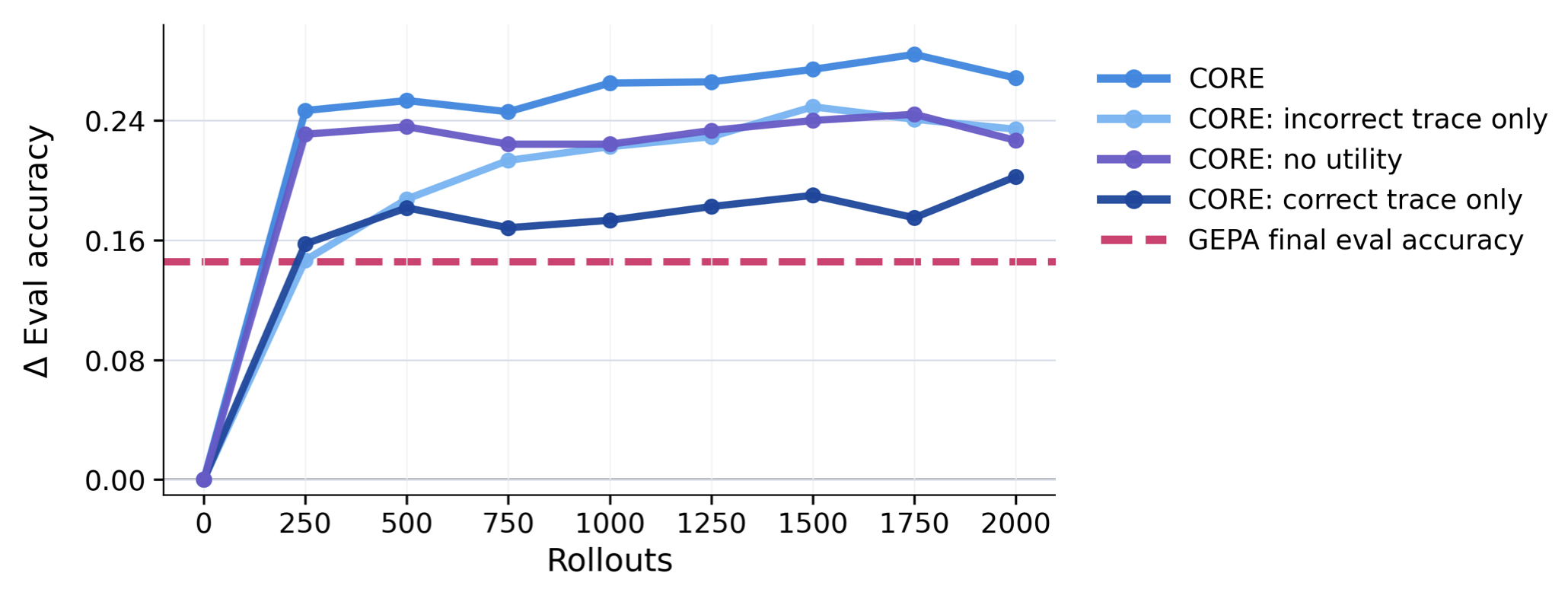}
\caption{
Ablations in the 10-training-example setting, aggregated across all four tasks. Y-axis shows improvement in held-out accuracy relative to rollout 0. \textit{Incorrect trace only} reflects only on the model's most recent incorrect reasoning trace, rather than comparing failure against success. \textit{Correct trace only} generates insights from a single successful trace, using a correct trace from the same problem when available. \textit{No utility} keeps contrastive insight generation but retrieves insights using only semantic similarity, without utility estimates. GEPA denotes the strongest non-CORE baseline.
}
\label{fig:ablations}
\end{figure}
\section{Discussion}

This work introduces CORE, a non-parametric learning algorithm that improves sample, rollout, and context efficiency over strong baselines across four logic, planning, and problem-solving tasks. The main implication is that learning from experience can be made more efficient by changing {\it what} is stored and reused. While standard episodic-memory methods store and retrieve entire rollouts or summaries of rollouts, CORE instead stores {\it insights} about rollouts: it contrasts successful and failed rollouts to produce candidate insights, gates these candidates through verifier feedback, and retrieves them based on both semantic relevance and observed utility. In this sense, CORE uses reflection as a means to perform credit assignment, with an insight being a hypothesis about which strategy or constraint to carry forward rather than a summary of what happened. Because these artifacts are inspectable and paired with empirical utility estimates, they are more transparent than weight updates and prompt optimization, with potential relevance to concerns about opaque self-improvement \citep{korbak2025cotmonitorability,emmons2025cotnecessary}. 

A natural extension is to combine CORE with RLVR-style training, so that non-parametric reflection can provide validated intermediate supervision for distillation. CORE's efficient use of added context and similarity-based retrieval also make continual learning a promising direction, since accumulating, merging, and selectively retrieving insights across tasks raises new challenges for learning from verifiable rewards. Another important direction is to extend CORE beyond single-turn reasoning problems to multi-step and agentic settings, where failures may occur at the level of plans, tool calls, subgoals, or environment interactions rather than final answers alone. Finally, CORE could be extended to multimodal domains by generating insights over paired visual and textual traces, allowing models to learn reusable constraints that connect perceptual evidence with text-based reasoning.

One important limitation of CORE is that it assumes access to verifiable rewards. This limits its applicability to verifiable domains. CORE's utility update also assigns the same outcome to all retrieved insights, making finer-grained credit assignment among multiple insights unresolved. Reflection and admission testing also introduce additional inference cost. Lastly, our experiments focus on reasoning, planning, and problem-solving tasks, which leaves open the question of how CORE performs in more open-ended environments.

\newpage
\section*{Acknowledgments}
The authors thank Andrew K. Lampinen, Giambattista Parascandolo, Kanishk Gandhi, and Jay McClelland for helpful feedback on this work. We also thank Thinking Machines Lab for providing research grant compute credits used to support experiments conducted using Tinker. J.E.F. is supported by NSF CAREER Award \#2436199, NSF DRL \#2400471, and awards from the Stanford Human-Centered AI Institute (HAI) and Stanford Accelerator for Learning. Finally, we thank MATS for contributing a grant for computational resources for research on methods in transparent reasoning.

\bibliographystyle{plainnat}
\bibliography{references}

@article{korbak2025cotmonitorability,
  title={Chain of thought monitorability: A new and fragile opportunity for ai safety},
  author={Korbak, Tomek and Balesni, Mikita and Barnes, Elizabeth and Bengio, Yoshua and Benton, Joe and Bloom, Joseph and Chen, Mark and Cooney, Alan and Dafoe, Allan and Dragan, Anca and others},
  journal={arXiv preprint arXiv:2507.11473},
  year={2025}
}

@article{emmons2025cotnecessary,
  title={When chain of thought is necessary, language models struggle to evade monitors},
  author={Emmons, Scott and Jenner, Erik and Elson, David K and Saurous, Rif A and Rajamanoharan, Senthooran and Chen, Heng and Shafkat, Irhum and Shah, Rohin},
  journal={arXiv preprint arXiv:2507.05246},
  year={2025}
}

@article{schwartz1998time,
  title={A time for telling},
  author={Schwartz, Daniel L and Bransford, John D},
  journal={Cognition and instruction},
  volume={16},
  number={4},
  pages={475--522},
  year={1998},
  publisher={Taylor \& Francis}
}

@article{alfieri2013learning,
  title={Learning through case comparisons: A meta-analytic review},
  author={Alfieri, Louis and Nokes-Malach, Timothy J and Schunn, Christian D},
  journal={Educational Psychologist},
  volume={48},
  number={2},
  pages={87--113},
  year={2013},
  publisher={Taylor \& Francis}
}

@article{cushman2020rationalization,
  title={Rationalization is rational},
  author={Cushman, Fiery},
  journal={Behavioral and Brain Sciences},
  volume={43},
  pages={e28},
  year={2020},
  publisher={Cambridge University Press}
}

@article{agrawal2025gepa,
  title={Gepa: Reflective prompt evolution can outperform reinforcement learning},
  author={Agrawal, Lakshya A and Tan, Shangyin and Soylu, Dilara and Ziems, Noah and Khare, Rishi and Opsahl-Ong, Krista and Singhvi, Arnav and Shandilya, Herumb and Ryan, Michael J and Jiang, Meng and others},
  journal={arXiv preprint arXiv:2507.19457},
  year={2025}
}

@article{guo2025deepseek,
  title={Deepseek-r1: Incentivizing reasoning capability in llms via reinforcement learning},
  author={Guo, Daya and Yang, Dejian and Zhang, Haowei and Song, Junxiao and Wang, Peiyi and Zhu, Qihao and Xu, Runxin and Zhang, Ruoyu and Ma, Shirong and Bi, Xiao and others},
  journal={arXiv preprint arXiv:2501.12948},
  year={2025}
}

@article{zelikman2022star,
  title={Star: Bootstrapping reasoning with reasoning},
  author={Zelikman, Eric and Wu, Yuhuai and Mu, Jesse and Goodman, Noah},
  journal={Advances in Neural Information Processing Systems},
  volume={35},
  pages={15476--15488},
  year={2022}
}

@article{shinn2023reflexion,
  title={Reflexion: Language agents with verbal reinforcement learning},
  author={Shinn, Noah and Cassano, Federico and Gopinath, Ashwin and Narasimhan, Karthik and Yao, Shunyu},
  journal={Advances in neural information processing systems},
  volume={36},
  pages={8634--8652},
  year={2023}
}

@article{didolkar2025metacognitive,
  title={Metacognitive reuse: Turning recurring llm reasoning into concise behaviors},
  author={Didolkar, Aniket and Ballas, Nicolas and Arora, Sanjeev and Goyal, Anirudh},
  journal={arXiv preprint arXiv:2509.13237},
  year={2025}
}

@article{lisman2005hippocampal,
  title={The hippocampal-VTA loop: controlling the entry of information into long-term memory},
  author={Lisman, John E and Grace, Anthony A},
  journal={Neuron},
  volume={46},
  number={5},
  pages={703--713},
  year={2005},
  publisher={Elsevier}
}

@article{adcock2006reward,
  title={Reward-motivated learning: mesolimbic activation precedes memory formation},
  author={Adcock, R Alison and Thangavel, Arul and Whitfield-Gabrieli, Susan and Knutson, Brian and Gabrieli, John DE},
  journal={Neuron},
  volume={50},
  number={3},
  pages={507--517},
  year={2006},
  publisher={Elsevier}
}

@article{nam2024systematic,
  title={Systematic human learning and generalization from a brief tutorial with explanatory feedback},
  author={Nam, Andrew J and McClelland, James L},
  journal={Open Mind},
  volume={8},
  pages={148--176},
  year={2024},
  publisher={MIT Press One Broadway, 12th Floor, Cambridge, Massachusetts 02142, USA~…}
}

@article{cobbe2021training,
  title={Training verifiers to solve math word problems},
  author={Cobbe, Karl and Kosaraju, Vineet and Bavarian, Mohammad and Chen, Mark and Jun, Heewoo and Kaiser, Lukasz and Plappert, Matthias and Tworek, Jerry and Hilton, Jacob and Nakano, Reiichiro and others},
  journal={arXiv preprint arXiv:2110.14168},
  year={2021}
}

@article{shao2024deepseekmath,
  title={Deepseekmath: Pushing the limits of mathematical reasoning in open language models},
  author={Shao, Zhihong and Wang, Peiyi and Zhu, Qihao and Xu, Runxin and Song, Junxiao and Bi, Xiao and Zhang, Haowei and Zhang, Mingchuan and Li, YK and Wu, Yang and others},
  journal={arXiv preprint arXiv:2402.03300},
  year={2024}
}

@article{wang2023voyager,
  title={Voyager: An open-ended embodied agent with large language models},
  author={Wang, Guanzhi and Xie, Yuqi and Jiang, Yunfan and Mandlekar, Ajay and Xiao, Chaowei and Zhu, Yuke and Fan, Linxi and Anandkumar, Anima},
  journal={arXiv preprint arXiv:2305.16291},
  year={2023}
}

@article{zhang2026memrl,
  title={Memrl: Self-evolving agents via runtime reinforcement learning on episodic memory},
  author={Zhang, Shengtao and Wang, Jiaqian and Zhou, Ruiwen and Liao, Junwei and Feng, Yuchen and Li, Zhuo and Zheng, Yujie and Zhang, Weinan and Wen, Ying and Li, Zhiyu and others},
  journal={arXiv preprint arXiv:2601.03192},
  year={2026}
}

@article{xu2026agent,
  title={Agent skills for large language models: Architecture, acquisition, security, and the path forward},
  author={Xu, Renjun and Yan, Yang},
  journal={arXiv preprint arXiv:2602.12430},
  year={2026}
}

@inproceedings{opsahl2024optimizing,
  title={Optimizing instructions and demonstrations for multi-stage language model programs},
  author={Opsahl-Ong, Krista and Ryan, Michael J and Purtell, Josh and Broman, David and Potts, Christopher and Zaharia, Matei and Khattab, Omar},
  booktitle={Proceedings of the 2024 Conference on Empirical Methods in Natural Language Processing},
  pages={9340--9366},
  year={2024}
}

@article{song2026expanding,
  title={Expanding the Capabilities of Reinforcement Learning via Text Feedback},
  author={Song, Yuda and Chen, Lili and Tajwar, Fahim and Munos, Remi and Pathak, Deepak and Bagnell, J Andrew and Singh, Aarti and Zanette, Andrea},
  journal={arXiv preprint arXiv:2602.02482},
  year={2026}
}

@inproceedings{lightman2023let,
  title={Let's verify step by step},
  author={Lightman, Hunter and Kosaraju, Vineet and Burda, Yuri and Edwards, Harrison and Baker, Bowen and Lee, Teddy and Leike, Jan and Schulman, John and Sutskever, Ilya and Cobbe, Karl},
  booktitle={The twelfth international conference on learning representations},
  year={2023}
}

@article{shi2026experiential,
  title={Experiential reinforcement learning},
  author={Shi, Taiwei and Chen, Sihao and Jiang, Bowen and Song, Linxin and Yang, Longqi and Zhao, Jieyu},
  journal={arXiv preprint arXiv:2602.13949},
  year={2026}
}

@article{shojaee2025illusion,
  title={The illusion of thinking: Understanding the strengths and limitations of reasoning models via the lens of problem complexity},
  author={Shojaee, Parshin and Mirzadeh, Iman and Alizadeh, Keivan and Horton, Maxwell and Bengio, Samy and Farajtabar, Mehrdad},
  journal={arXiv preprint arXiv:2506.06941},
  year={2025}
}

@article{opedal2024mathgap,
  title={MathGAP: Out-of-distribution evaluation on problems with arbitrarily complex proofs},
  author={Opedal, Andreas and Shirakami, Haruki and Sch{\"o}lkopf, Bernhard and Saparov, Abulhair and Sachan, Mrinmaya},
  journal={arXiv preprint arXiv:2410.13502},
  year={2024}
}

@article{lin2025zebralogic,
  title={Zebralogic: On the scaling limits of llms for logical reasoning},
  author={Lin, Bill Yuchen and Bras, Ronan Le and Richardson, Kyle and Sabharwal, Ashish and Poovendran, Radha and Clark, Peter and Choi, Yejin},
  journal={arXiv preprint arXiv:2502.01100},
  year={2025}
}

@article{ollinger2008investigating,
  title={Investigating the effect of mental set on insight problem solving},
  author={{\"O}llinger, Michael and Jones, Gary and Knoblich, G{\"u}nther},
  journal={Experimental psychology},
  volume={55},
  number={4},
  pages={269--282},
  year={2008},
  publisher={Hogrefe \& Huber Publishers}
}

@article{knoblich1999constraint,
  title={Constraint relaxation and chunk decomposition in insight problem solving.},
  author={Knoblich, G{\"u}nther and Ohlsson, Stellan and Haider, Hilde and Rhenius, Detlef},
  journal={Journal of Experimental Psychology: Learning, memory, and cognition},
  volume={25},
  number={6},
  pages={1534},
  year={1999},
  publisher={American Psychological Association}
}

@article{ouyang2025reasoningbank,
  title={Reasoningbank: Scaling agent self-evolving with reasoning memory},
  author={Ouyang, Siru and Yan, Jun and Hsu, I and Chen, Yanfei and Jiang, Ke and Wang, Zifeng and Han, Rujun and Le, Long T and Daruki, Samira and Tang, Xiangru and others},
  journal={arXiv preprint arXiv:2509.25140},
  year={2025}
}

@article{agarwal2025gpt,
  title={gpt-oss-120b \& gpt-oss-20b model card},
  author={Agarwal, Sandhini and Ahmad, Lama and Ai, Jason and Altman, Sam and Applebaum, Andy and Arbus, Edwin and Arora, Rahul K and Bai, Yu and Baker, Bowen and Bao, Haiming and others},
  journal={arXiv preprint arXiv:2508.10925},
  year={2025}
}

@article{gunther2023jina,
  title={Jina embeddings 2: 8192-token general-purpose text embeddings for long documents},
  author={G{\"u}nther, Michael and Ong, Jackmin and Mohr, Isabelle and Abdessalem, Alaeddine and Abel, Tanguy and Akram, Mohammad Kalim and Guzman, Susana and Mastrapas, Georgios and Sturua, Saba and Wang, Bo and others},
  journal={arXiv preprint arXiv:2310.19923},
  year={2023}
}

@misc{tml2025tinker,
  author = {Thinking Machines Lab},
  title = {Tinker},
  year = {2025},
  url = {https://thinkingmachines.ai/tinker/},
}

@article{nasvytis2026leveraging,
  title={Leveraging Speech to Identify Signatures of Insight and Transfer in Problem Solving},
  author={Nasvytis, Linas and Fan, Judith E},
  journal={arXiv preprint arXiv:2605.12970},
  year={2026}
}

@article{zhang2026useful,
  title={Useful Memories Become Faulty When Continuously Updated by LLMs},
  author={Zhang, Dylan and Lin, Yanshan and Wu, Zhengkun and Sun, Yihang and Li, Bingxuan and Li, Dianqi and Peng, Hao},
  journal={arXiv preprint arXiv:2605.12978},
  year={2026}
}

@article{mcclelland1995there,
  title={Why there are complementary learning systems in the hippocampus and neocortex: insights from the successes and failures of connectionist models of learning and memory.},
  author={McClelland, James L and McNaughton, Bruce L and O'Reilly, Randall C},
  journal={Psychological review},
  volume={102},
  number={3},
  pages={419},
  year={1995},
  publisher={American Psychological Association}
}

\newpage
\appendix
\section*{Appendix Outline}
\begin{itemize}
    \item Pseudocode for the CORE algorithm.
    \item Growth of the insight memory across training rollouts.
    \item Distributions of insight utilities.
\end{itemize}

\section{Pseudocode for the CORE algorithm}\label{app:pseudocode}
Algorithm~\ref{alg:core} gives pseudocode for the CORE training procedure, including failure-biased sampling, insight retrieval, contrastive reflection, and admission testing.

\begin{algorithm}[!htbp]
\caption{CORE Training}
\label{alg:core}
\small
\begin{algorithmic}[1]
\Require Training set $\mathcal D_{\mathrm{train}}$, frozen model $M$, verifiers $\{V_q\}$, rollout memory $\mathcal R$, insight memory $\mathcal I$
\State Initialize $\mathcal R \gets \emptyset$, $\mathcal I \gets \emptyset$
\State Estimate initial no-memory baselines $b_q$ for each $q \in \mathcal D_{\mathrm{train}}$
\For{$t = 1,\ldots,T_{\mathrm{train}}$}
    \State Sample training problem $q$ using failure-biased sampling
    \State Retrieve top-$K$ insights $L$ from $\mathcal I$ using neighbors of $q$ in $\mathcal R$
    \State Generate solution $y \gets M(q, L)$ and reward $r \gets V_q(y)$
    \State Compute utility $U(q) \gets r - b_q$
    \State Update insight-memory statistics for each $\ell \in L$ using $U(q)$
    \State Store rollout $\tau=(q,L,y,r)$ in rollout memory $\mathcal R$
    \If{$r=0$}
        \State Retrieve positive rollout $\tau^+$ from $\mathcal R$
        \If{$\tau^+$ exists}
            \State Generate candidate insights $C \gets M(\tau,\tau^+)$
            \State Filter candidate insights $C$
            \ForAll{$\ell \in C$}
                \State Admission-test $\ell$ on $q$
                \If{$\ell$ passes admission}
                    \State Add $\ell$ to insight memory $\mathcal I$
                    \State Initialize utility statistics for $\ell$
                \EndIf
            \EndFor
        \EndIf
    \EndIf
\EndFor
\end{algorithmic}
\end{algorithm}

\section{Growth of the insight memory across training rollouts}
\begin{figure}[t]
  \centering
\includegraphics[width=0.8\linewidth]{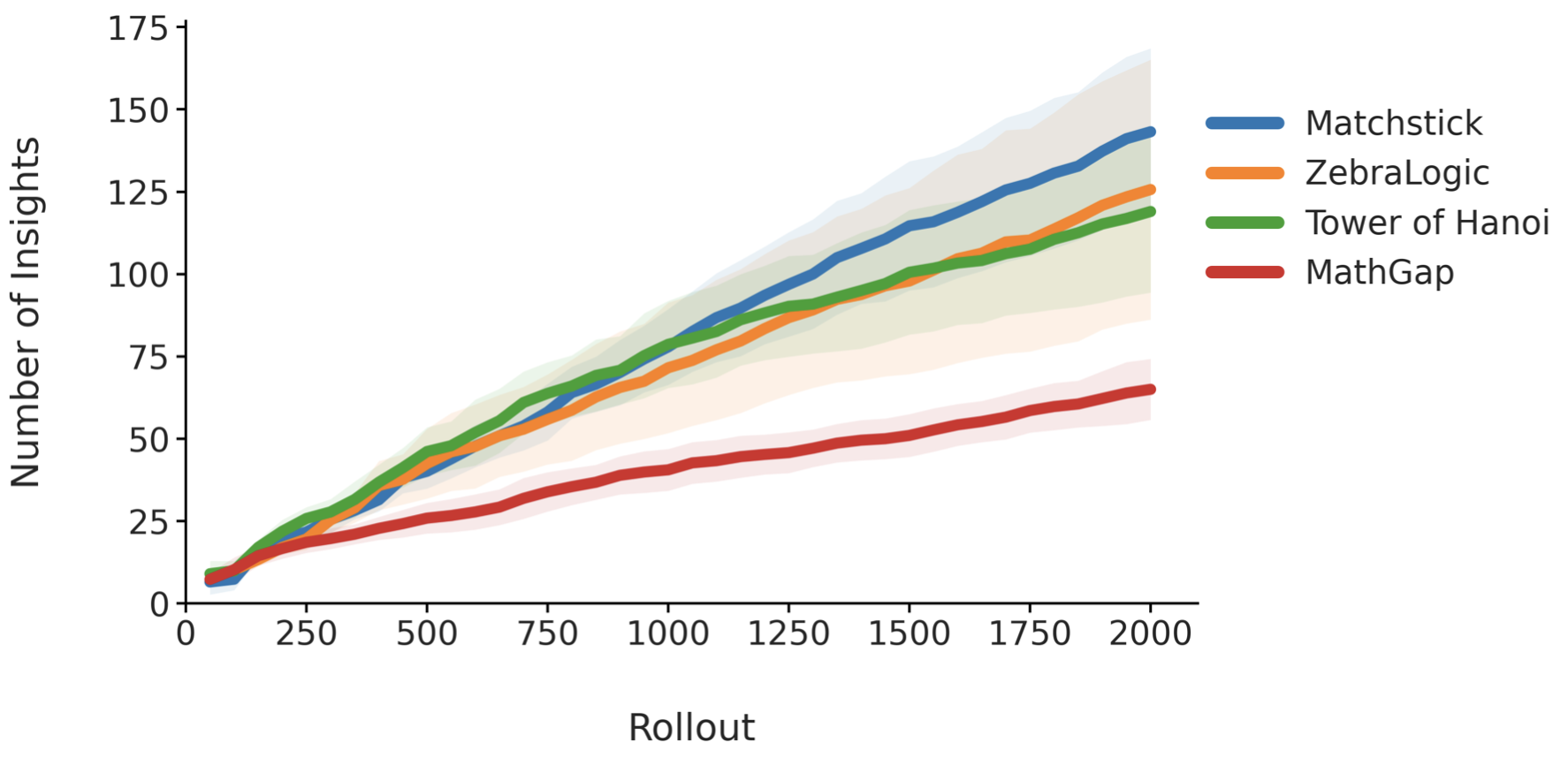}
  \caption{Growth of the insight memory across rollouts for each task. Lines show the number of stored insights accumulated during training. Ribbons show 95\% confidence intervals across the runs.}
  \label{fig:memory_growth}
\end{figure}

Figure~\ref{fig:memory_growth} shows how the number of admitted insights grows across training rollouts for each task. Ribbons show 95\% confidence intervals across the runs.

\section{Distributions of insight utilities}
\begin{figure}[t]
  \centering
\includegraphics[width=1.0\linewidth]{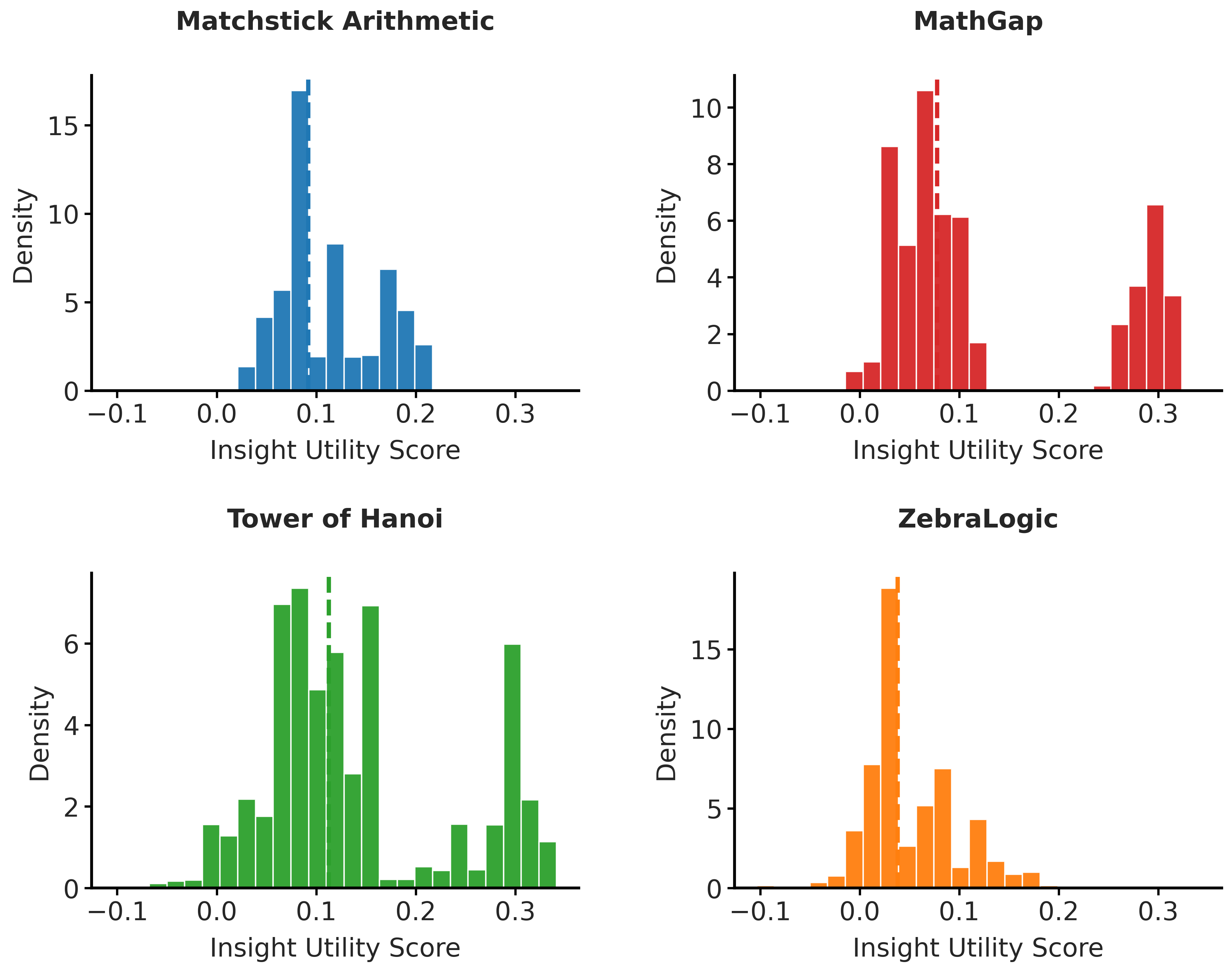}
  \caption{Distributions of insight utilities across the four tasks.}
  \label{fig:utility_histograms}
\end{figure}

Figure~\ref{fig:utility_histograms} shows the distribution of estimated insight utilities across the four tasks.

\end{document}